\pdfoutput=1

\documentclass[11pt]{article}

\usepackage{EACL2023}
\usepackage{comment}
\usepackage{times}
\usepackage{hyperref}
\usepackage[T1]{fontenc}
\usepackage{graphics}
\usepackage{graphicx}
\usepackage{tabularx}
\usepackage{enumitem}

\usepackage[utf8]{inputenc}

\newcommand{\primeqa}{\textsc{PrimeQA}}

\usepackage{microtype}

\usepackage{inconsolata}
\usepackage{graphics}
\usepackage{xparse}
\usepackage{booktabs}
\usepackage{subfig}

\NewDocumentCommand{\vk}
{ mO{} }{\textcolor{magenta}{\textsuperscript{\textit{vk}}\textsf{\textbf{\small[#1]}}}}
\NewDocumentCommand{\jaydeep}
{ mO{} }{\textcolor{violet}{\textsuperscript{\textit{Jaydeep}}\textsf{\textbf{\small[#1]}}}}
\NewDocumentCommand{\avi}
{ mO{} }{\textcolor{red}{\textsuperscript{\textit{Avi}}\textsf{\textbf{\small[#1]}}}}
\NewDocumentCommand{\arafat}
{ mO{} }{\textcolor{blue}{\textsuperscript{\textit{Arafat}}\textsf{\textbf{\small[#1]}}}}
\NewDocumentCommand{\mihaela}
{ mO{} }{\textcolor{green}{\textsuperscript{\textit{Mihaela}}\textsf{\textbf{\small[#1]}}}}
\NewDocumentCommand{\sara}
{ mO{} }{\textcolor{orange}{\textsuperscript{\textit{Sara}}\textsf{\textbf{\small[#1]}}}}
\NewDocumentCommand{\scott}
{ mO{} }{\textcolor{olive}{\textsuperscript{\textit{Scott}}\textsf{\textbf{\small[#1]}}}}

\NewDocumentCommand{\rbhat}
{ mO{} }{\textcolor{brown}{\textsuperscript{\textit{Rbhat}}\textsf{\textbf{\small[#1]}}}}

\NewDocumentCommand{\martin}
{ mO{} }{\textcolor{cyan}{\textsuperscript{\textit{Martin}}\textsf{\textbf{\small[#1]}}}}

\NewDocumentCommand{\bhavani}
{ mO{} }{\textcolor{purple}{\textsuperscript{\textit{Bhavani}}\textsf{\textbf{\small[#1]}}}}

\title{\primeqa{}: The Prime Repository for State-of-the-Art Multilingual Question Answering Research and Development}

\author{Avirup Sil\thanks{\ \ Corresponding author}, Jaydeep Sen,  Bhavani Iyer, Martin Franz, Kshitij Fadnis, \\
 \textbf{Mihaela Bornea, Sara Rosenthal, Scott McCarley, Rong Zhang, Vishwajeet Kumar,}\\ \textbf{Yulong Li, Md Arafat Sultan,  Riyaz Bhat, Radu Florian, Salim Roukos}\\ 
IBM Research AI}

\usepackage{color}

\newcommand{\eat}[1]{} 

\eat{}

\begin{document}

\maketitle
\begin{abstract}
The field of Question Answering (QA) has made remarkable progress in recent years, thanks to the advent of large pre-trained language models, newer realistic benchmark datasets with leaderboards, and novel algorithms for key components such as retrievers and readers.
In this paper, we introduce \primeqa{}: a one-stop and open-source QA repository with an aim to democratize QA research and facilitate easy replication of state-of-the-art (SOTA) QA methods.
\primeqa{} supports core QA functionalities like retrieval and reading comprehension as well as auxiliary capabilities such as question generation.
It has been designed as an end-to-end toolkit for various use cases: building front-end applications, replicating SOTA methods on public benchmarks, and expanding pre-existing methods.
\primeqa{} is available at: \url{https://github.com/primeqa}.
\end{abstract}

\section{Introduction}
\label{sec:intro}
Question Answering (QA) is 
a major area of interest in Natural Language Processing (NLP), consisting primarily of two subtasks:
information retrieval (IR) \cite{manning2008introduction,schutze2008introduction} and machine reading comprehension (MRC) \cite{rajpurkar-etal-2016-squad,rajpurkar-etal-2018-know,kwiatkowski-etal-2019-natural,chakravarti2020towards}. 
IR and MRC systems, also referred to as \textit{retrievers} and \textit{readers}, respectively, are commonly assembled in an end-to-end open-retrieval QA pipeline (henceforth, OpenQA) \cite{chen-etal-2017-reading, lee-etal-2019-latent, karpukhin-etal-2020-dense, santhanam-etal-2022-colbertv2} that accepts a query and a large document collection as its input and provides an answer as output.
The specific role of the retriever is to identify documents or passages (i.e., \textit{contexts}) that contain information relevant to the query, while the reader component extracts a precise answer from such contexts. 

While QA as a field
has advanced rapidly, 
software
to perform and replicate QA experiments 
has mostly been written
in silos. 
At the time of this writing, no central repository exists that facilitates the training, analysis and augmentation of state-of-the-art (SOTA) models for different QA tasks at scale.
In view of the above, and with an aim to democratize QA research by providing easy replicability, here, we propose \primeqa{}: 
an open-source repository\footnote{
\url{https://github.com/primeqa}
} designed as an end-to-end toolkit, with 
all the necessary tools to 
easily and quickly
create a custom QA application. We provide a main repository that contains easy-to-use scripts 
for
retrieval, machine reading comprehension, and question generation with the ability to perform training, inference, and 
performance evaluation.
Additionally, several sibling repositories provide features for easily connecting various retrievers and readers and creating a front-end user interface (UI) for end users. \primeqa{} has been designed as a platform for QA development and research and encourages collaboration from everyone in the field from beginners to experts. \primeqa{} already has a growing developer base with contributions from major academic institutions. 

Our paper makes several major contributions:
\begin{itemize}\setlength\itemsep{-0.1em}
\item We present \primeqa{}, 
a first-of-its-kind repository for
comprehensive QA research. It is free to use, well documented, easy to contribute to, and license friendly (Apache 2.0) for both academic and commercial usage.
\item \primeqa{} provides the mechanism via accompanying repositories to create custom OpenQA applications containing both retrievers and readers for industrial deployment including a front-end UI.
\item We provide \textit{easy-to-use} implementations of SOTA
retrievers and readers that are at the top of major QA leaderboards, 
with capabilities for performing training, inference and performance evaluation of these models.
\item \primeqa{} models are built on top of Transformers \cite{wolf-etal-2020-transformers} and are available on the Hugging Face model hub.\footnote{\url{https://huggingface.co/PrimeQA}} 
\end{itemize}

\section{Related Work}
\label{sec:related-work}

One of the largest open source community efforts for NLP software is Papers with Code \cite{papers-with-code}. Their mission is to create a free and open resource for NLP papers, code, datasets, methods and evaluation tables. Their focus is to cater to the wider NLP and Machine Learning community and not just QA. Even though the QA section includes over 1800 papers with their code, the underlying software components (written in various versions of both pytorch and tensorflow, with no central control whatsoever) do not communicate with each other. These disjoint QA resources hinder replicability and effective collaboration, and ultimately lead to quick sunsetting of new capabilities.

Recently, among the most used repositories for NLP users have been the Transformers repository \cite{wolf-etal-2020-transformers}. 
However, while being widely adopted by the community, they lack a distinct focus on QA. Unlike \primeqa{}, they only support \href{https://github.com/huggingface/transformers/blob/main/examples/pytorch/question-answering/run_qa.py}{one general script} for extractive QA and several stand-alone python scripts for retrievers. 
Similarly FairSeq, \cite{ott2019fairseq} and AllenNLP \cite{gardner2018allennlp} also focus on a wide array of generic NLP tasks and hence do not solely present a QA repository allowing anyone plug and play components for their custom search application. There exists several toolkits catered to build customer-specific search applications \cite{nvidia-tao, haystack} or search-based virtual assistants \cite{watson-assistant}. However, while they have a good foundation for software deployment, unlike \primeqa{}, they lack the focus on replicating (and extending) the latest SOTA for QA on academic benchmarks which is an essential component needed in order for us to make rapid progress in this field.

\section {
\primeqa{}}
\begin{figure*}[ht]
\centering
 \includegraphics[width=2\columnwidth]{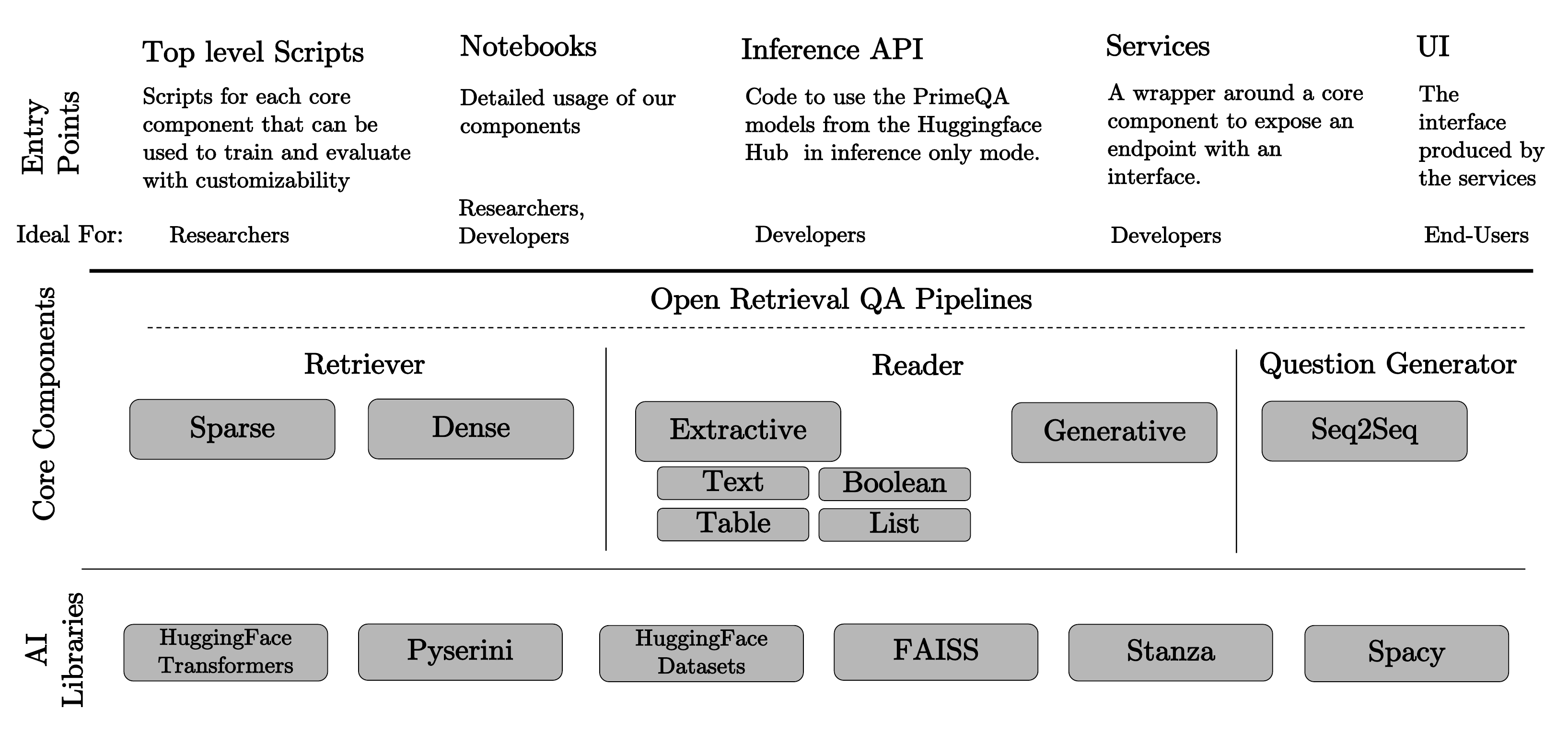}
 \caption{The \primeqa{} Repository: the core components and features.}
 \label{fig:pqa_tree}     
\end{figure*}

\begin{table}[t]\small								
\centering								
\begin{tabularx}{\columnwidth}{ll}
\toprule								
\textbf{Core Models}	&	\textbf{Extensions}				\\		
\midrule								
              \multicolumn{2}{c}								
              {\textbf{Retriever}}		\\						
\midrule								
BM25{\scriptsize ~\cite{RobertsonBM25}}	&	Dr.DECR	\href{https://nlp.cs.washington.edu/xorqa/}{*} {\scriptsize ~\cite{li-etal-2022-learning-cross}} 			\\	
DPR	{\scriptsize ~\cite{karpukhin-etal-2020-dense}} &						\\	
ColBERT	{\scriptsize ~\cite{santhanam-etal-2022-colbertv2}} &						\\	
\midrule								
             \multicolumn{2}{c}{\textbf{Reader}}								
             \\								
 \midrule								
General MRC {\scriptsize ~\cite{alberti2019bert}}	&	ReasonBERT				\\		
FiD	{\scriptsize ~\cite{FID}} &	OmniTab{\scriptsize ~\cite{jiang2022omnitab}}					\\		
Boolean\href{https://ai.google.com/research/tydiqa}{*} {\scriptsize ~\cite{https://doi.org/10.48550/arxiv.2206.08441}} 										\\				
	Lists 	\\
Tapas{\scriptsize ~\cite{herzig2020tapas}}					\\	
Tapex{\scriptsize ~\cite{liu2021tapex}}	&  \\	

\midrule								
\multicolumn{2}{c}{\textbf{Question Generation}}		\\						
\midrule								
Table QG {\scriptsize ~\cite{t3qa}}					\\	
Passage QG	&						\\	
Hybrid QG	&						\\	
 \bottomrule								
\end{tabularx}								
\caption{ A non-exhaustive list of core \primeqa{} models for the three main supported tasks (left) and their various extensions (right) available on our Hugging Face model hub: \url{https://huggingface.co/PrimeQA}. * SOTA leaderboard systems.
}								
\label{tab:models_apps_tasks}								
\end{table}	

\primeqa{} is a comprehensive open-source resource for cutting-edge QA research and development, governed by the following design principles:

\noindent$\bullet$ \textbf{Reproducible:} 
Users can easily reproduce SOTA published results
and combine or ``plug and play'' the corresponding methods with companion modules, e.g., a reader with a retriever.
The various supported applications and the associated \primeqa{} components are listed in Table~\ref{tab:models_apps_tasks}.

\noindent$\bullet$ \textbf{Customizable:} We allow users to customize and extend SOTA models for their own applications. 
This often entails fine-tuning on users' custom data, which they can provide through one of several supported data formats, or process on their own by writing a custom data processor.

\noindent$\bullet$ \textbf{Reusable:} 
We aim to make it straightforward for developers to quickly deploy pre-trained off-the-shelf \primeqa{} models for their 
QA
applications, requiring minimal code change.

\noindent$\bullet$ \textbf{Accessible:} We provide easy integration with 
Hugging Face Datasets and the Model Hub, allowing users to quickly plug in 
a range of
datasets and models as shown in Table~\ref{tab:models_apps_tasks}.

\primeqa{} 
in its entirety is a collection of
four different repositories: 
a primary \textit{research and replicability}\footnote{
\scriptsize \url{https://github.com/primeqa/primeqa}
} repository and three accompanying repositories\footnote{
\scriptsize \url{https://github.com/primeqa/create-primeqa-app}
}$^{,}$\footnote{
\scriptsize \url{https://github.com/primeqa/primeqa-orchestrator}
}$^{,}$\footnote{
\scriptsize \url{https://github.com/primeqa/primeqa-ui}
} for industrial deployment. Figure~\ref{fig:pqa_tree} shows a diagram of the  PrimeQA repository. 
It 
provides
several 
entry points, supporting
the needs of different users, 
as shown at the top of the figure. The repository is centered around three core components: 
a \textbf{retriever}, a \textbf{reader}, and a \textbf{question generator} for data augmentation.
These components can be used as individual modules or assembled into an end-to-end QA pipeline. 
All components are implemented
on top of existing AI libraries. 

\subsection{
The Core Components
} 


Each of the three core \primeqa{} components supports different flavors of the task it has been built for, as we detail in this section.


\subsubsection{Retriever: \texttt{run\_ir.py}}
Retrievers predict documents (or passages) from a document collection that are relevant to an input question.  
\primeqa{}  has both sparse and SOTA dense retrievers along with their extensions, as shown in  Table~\ref{tab:models_apps_tasks}.
We provide a
single 
Python
script \texttt{run\_ir.py} that
can be passed arguments to switch between different retriever algorithms.\\
\noindent\textbf{Sparse:} 
BM25 \cite{RobertsonBM25} is one of the most popular sparse retrieval methods, thanks to its simplicity, efficiency and robustness.
Our Python-based implementation of BM25 is powered by the open-source library \href{https://github.com/castorini/pyserini}{PySerini}.

\noindent\textbf{Dense:}  
Modern neural retrievers have utilized dense question and passage representations to achieve SOTA performance
on various benchmarks, 
while needing GPUs for efficiency.  We currently support ColBERT \cite{santhanam-etal-2022-colbertv2} and DPR \cite{karpukhin-etal-2020-dense}: both  
fine-tune pre-trained language models to train question and passage encoders \cite{devlin-etal-2019-bert, conneau-etal-2020-unsupervised}.
They utilize FAISS \cite{faiss2017} for 
K-nearest neighbor clustering and compressed index representations, respectively, 
and support multilingual retrieval with 
the question and 
the documents
being in the same \cite{lee-etal-2019-latent,longpre2021mkqa} or different languages (cross-lingual) 
\cite{asai-etal-2021-xor}.

\subsubsection{Reader: \texttt{run\_mrc.py}}
Given a question and a 
retrieved passage---also called the \textit{context}---a reader 
predicts an answer that is either extracted directly from the context or is generated based on it.
\primeqa{} supports training and inference of both extractive and generative readers  through a single Python script: \texttt{run\_mrc.py}.
It works out-of-the-box with 
different
QA models extended from the Transformers 
library \cite{wolf-etal-2020-transformers}.

\noindent\textbf{Extractive:} 
\primeqa{}'s
general extractive reader is a pointer network that predicts the start and end of the answer 
span
from the input context \cite{devlin-etal-2019-bert,alberti2019bert}. 
It can be initialized with most large pre-trained language models \cite{devlin-etal-2019-bert, roberta, conneau-etal-2020-unsupervised}. In addition, our reader is extremely versatile as it
can
provide responses to questions 
with list answers \cite{khashabi-etal-2021-gooaq-open}, 
\textit{yes/no} responses
to 
Boolean
questions \cite{clark-etal-2019-boolq, clark-etal-2020-tydi, naturalQuestions}, answer spans found in 
tables \cite{herzig-etal-2020-tapas} and in multimodal (text+image) documents \cite{mathew2021docvqa}. 
Examples of several extractive readers along with their extensions are 
provided
in Table~\ref{tab:models_apps_tasks}. 


\noindent\textbf{Generative:} 
\primeqa{} provides generative readers
based on the popular Fusion-in-Decoder (FiD) \cite{FID} algorithm. Currently, it supports easy initialization with large pre-trained sequence-to-sequence (henceforth, seq2seq) models \cite{BART,T5}. With FiD, the question and the retrieved passages are used to generate 
relatively
long and complex multi-sentence answers providing support for long form question answering tasks, \textit{e.g.}, ELI5 \cite{kilt-eli5, eli5}. 

\subsubsection{Question Generation: \texttt{run\_qg.py}}
Data augmentation through synthetic question generation (QG) can be a powerful tool for improving QA model generalization~ \cite{ alberti-etal-2019-synthetic,sultan-etal-2020-importance,reddy2022entity}, including in   domain adaptation \cite{shakeri-etal-2021-towards,gangi2021synthetic,gangi-reddy-etal-2022-towards}, domain generalization \cite{sultan2022not} and few-shot learning \cite{yue2022qve} settings. QG can also circumvent the problem of not having labeled data in the target domain of application. Question generators take a span of text (e.g., a  sentence) from a document as input, hypothesize an answer (e.g., a named entity) and generate a question as output. \primeqa{}'s QG component ~\cite{t3qa} is based on SOTA sequence-to-sequence generation architectures ~\cite{T5}, and supports both unstructured and structured input text through a single Python script \texttt{run\_qg.py}. 
A list of available question generators and their extensions are provided in Table~\ref{tab:models_apps_tasks}.\\
\noindent\textbf{Unstructured Input:} 
Our first variant of QG is a multilingual text-to-text model capable of generating questions in the language of the input passage. It fine-tunes a pre-trained T5 language model \cite{T5} on publicly available multilingual QA data ~\cite{tydiqa}. 

\noindent\textbf{Structured Input:} We also provide QG capability over tables, for which the generator is trained on examples of SQL and natural language question pairs extracted from the popular Table QA dataset ~\cite{wikisql}. As in ~\cite{t3qa},
during inference, \primeqa{} uses a controllable SQL sampler to select SQL queries for a given table and answer text, and then applies the trained QG model to generate natural language questions.

\subsection{Entry Points}

We cater to different user groups in the QA community by providing different entry points to \primeqa{}, as shown in Figure~\ref{fig:pqa_tree}.

\noindent$\bullet$ \textbf{Top-level Scripts:}  Researchers can use the top level scripts, run\_ir/mrc/qg.py, to reproduce published results as well as train, fine-tune and evaluate associated models on their own custom data.

\noindent$\bullet$ \textbf{Jupyter Notebooks:}  These demonstrate how to use built-in classes to 
run the different \primeqa{} components and perform the corresponding tasks. These are useful for developers and  researchers who want to reuse and extend \primeqa{} functionalities.

\noindent$\bullet$ \textbf{Inference APIs:} The Inference APIs are primarily meant for developers, allowing them to use \primeqa{} components on their own data with only a few lines of code. These APIs can be initialized with the pre-trained \primeqa{} models provided in the HuggingFace hub, or with a custom model that has been trained for a specific use case.  

\noindent$\bullet$ \textbf{Service Layer:} The service layer helps developers set up an end-to-end QA system quickly by providing a wrapper around the core components that exposes an endpoint and an API. 

\noindent$\bullet$ \textbf{UI:} The UI is for end-users, including the non-technical layman who wants to use \primeqa{} services interactively to ask questions and get answers.

\subsection{Pipelines for OpenQA}
\primeqa{} core components and entry points make it intuitive for users to build an OpenQA \textit{pipeline} and configure it to use any of the \primeqa{} retrievers and readers.
This is facilitated through a lightweight wrapper built around each core component, which implements a training and an inference API.
The retrieval component of the pipeline predicts relevant passages/contexts for an input question, and the reader predicts an answer from the retrieved contexts. \primeqa{} pipelines are easy to construct using the pre-trained models in the model hub and our inference APIs. 

An example of such a pipeline can be connecting a ColBERT retriever to an FiD reader to construct a long form QA (LFQA) system. This pipeline uses the retriever to obtain supporting passages that are subsequently used by the reader to generate complex multi-sentence answers. A different pipeline can also be instantiated to use an extractive reader instead that is available through our model hub.




\section{Services and Deployment}
\label{sec:services}

\begin{figure*}[!ht]
 \includegraphics[width=2\columnwidth]{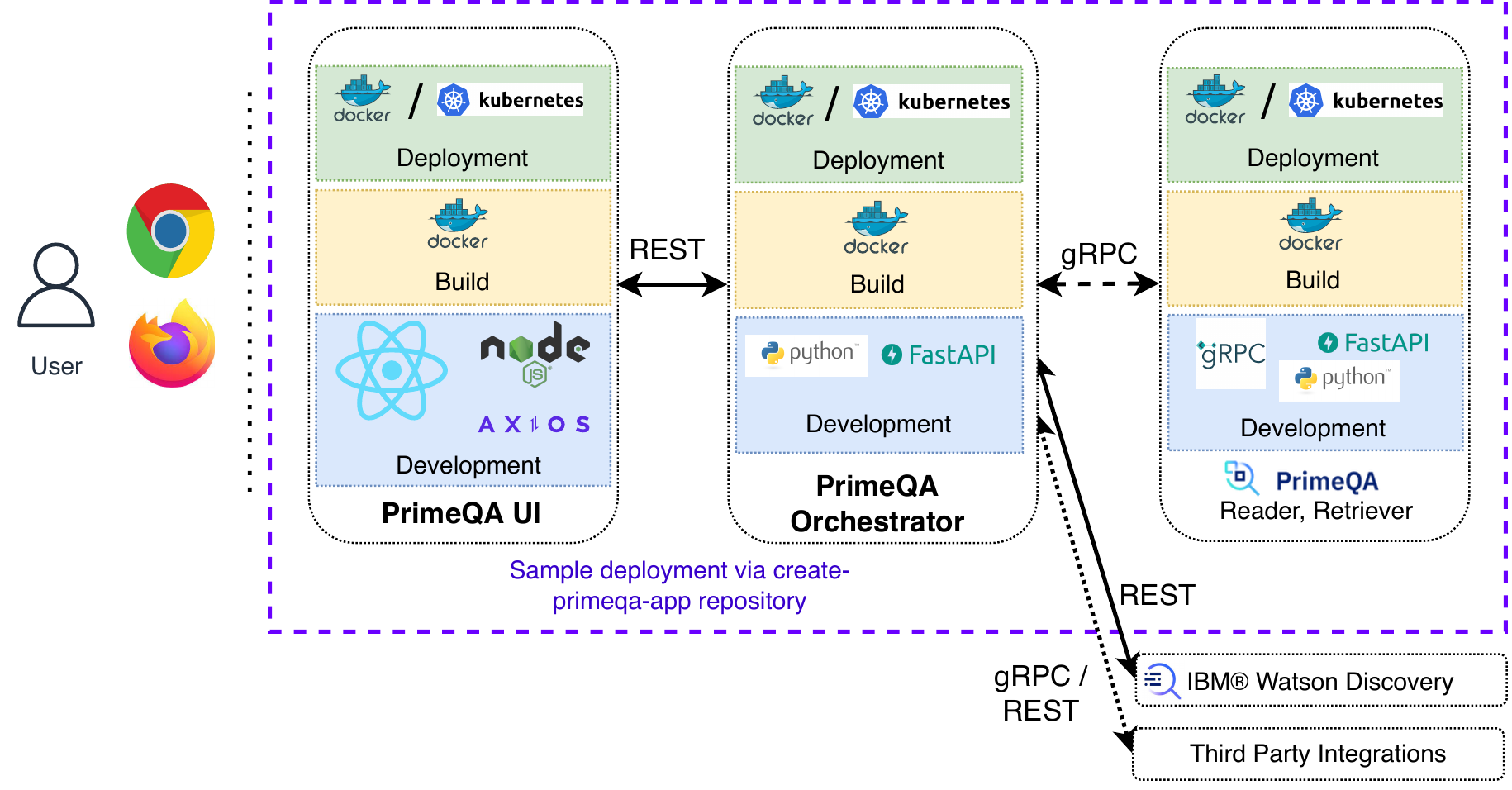}
 \caption{\primeqa{}'s end-to-end application. Each container contains a development (blue), build (yellow) and deployment (green) stack.}
 \label{fig:services_arch}     
\end{figure*}

\eat{
While research seeks to operate in a more bleeding edge environment, industrial deployment often necessitates running models at scale. More importantly, while a reader component can certainly be written with one single programming language, e.g. python, a custom search application may need to interact with a retriever written in another programming language,  which again may need to interact with other components for data ingestion, data preparation, front-end services and etc. Docker allows users to better manage all these interactions as each OpenQA ``micro-service'' may be written in different languages allowing scalability and easy addition or deletion of independent services. \primeqa{} brings reproducibility, portability, easy deployment, granular updates and simplicity via its services available on the public Docker hub \footnote{
\url{https://hub.docker.com/u/primeqa}
}. This \textit{service layer} allows \primeqa{} users to easily reproduce the working environment that is used to train and run the OpenQA model anywhere. It enables the packaging of code and dependencies into containers that can be ported to different servers even if it is a different hardware or operating system with additional resources such as GPUs, more memory or powerful CPUs. We make it easy to deploy and make our models available by wrapping them into APIs in a container and deploying the container using technology such as \avi{to dos} OpenShift, a Kubernetes distribution. 
}

Industrial deployment often necessitates running complex models and processes at scale. 
We use Docker to package these components into micro-services that interact with each other and can be ported to servers with different hardware capabilities (e.g. GPUs, CPUs, memory).
The use of docker makes the addition, replacement or deletion of services easy and scalable.
\eat{For example, the reader can interact with a retriever written in another programming language. The retriever also interacts with other components for data ingestion, data preparation, front-end services, etc.}
All components in the \primeqa{} repository are available via REST and/or gRPC micro-services.
Our docker containers are available on the public \href{https://hub.docker.com/u/primeqa}{\texttt{DockerHub}} and can be deployed using technologies such as OpenShift and Kubernetes. 

In addition to the main  \href{https://github.com/primeqa/primeqa}{\texttt{\primeqa{}}} repository, we provide three sibling repositories for application deployment:
\begin{description}[noitemsep]
\item \href{https://github.com/primeqa/primeqa-ui}{\texttt{primeqa-ui}} is the front-end UI. Users can personalize the front-end UI by adding custom organization logos or changing the display fonts.
\item \href{https://github.com/primeqa/primeqa-orchestrator}{\texttt{primeqa-orchestrator}}  is a REST server and is the central hub for the integration of \primeqa{} services and external components and the execution of a pipeline. For instance, the orchestrator can be configured to search a document collection with either a retriever from PrimeQA such as ColBERT, or an external search engine such as Watson Discovery.\footnote{
\url{https://www.ibm.com/cloud/watson-discovery}}
\item \href{https://github.com/primeqa/create-primeqa-app}{\texttt{create-primeqa-app}} provides the scripts to launch the demo application by starting the orchestrator and UI services.
\end{description}

\begin{figure*}[!ht]\centering
  \fbox{{\includegraphics[trim=0 100 0 120,clip,width=2\columnwidth]{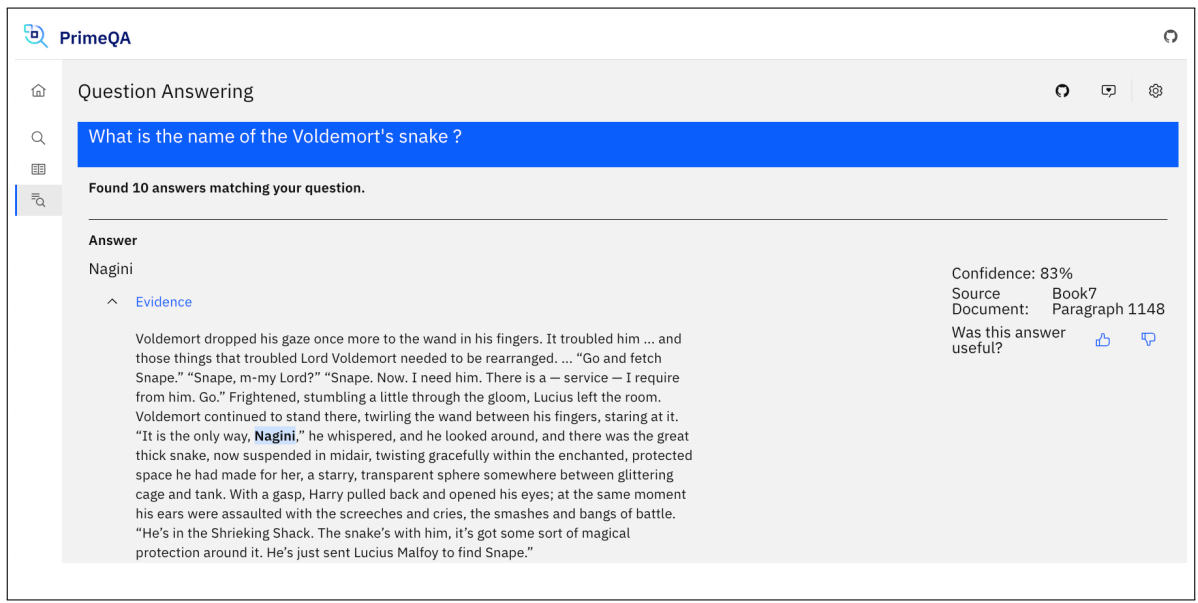}}}
\caption{A custom OpenQA search application built with \primeqa{}. Additional screenshots are in Appendix~\ref{sec:appendix}.}
\label{fig:primeqa-app-demo-qa} 
\end{figure*}

Figure~\ref{fig:services_arch} illustrates how to deploy a QA application at scale using the core PrimeQA services (e.g. Reader and Retriever) and our three sibling repositories. We provide this end-to-end deployment for our demo, however users can also utilize PrimeQA as an application with their own orchestrator or UI. 

Figure~\ref{fig:primeqa-app-demo-qa} shows an OpenQA demo application built with \primeqa{} components.
Our demo application provides a mechanism to collect user feedback. The \textit{thumbs up / down} icons next to each result enables a user to record feedback which is then stored in a database. The user feedback data can then be retrieved and used as additional training data to further improve a retriever and reader model.

\section{Community Contributions}
\label{sec:community}
While being relatively new, \primeqa{} has already garnered positive attention from the QA community and is receiving constant successful contributions from both international academia and industry via Github pull requests. We describe some instances here and encourage further contributions from all in the community. We provide support for those interested in contributing through a dedicated slack channel \footnote{\href{https://ibm.biz/pqa-slack}{https://ibm.biz/pqa-slack}}, Github issues and PR reviews. 


\noindent\textbf{Neural Retrievers:} ColBERT, one of our core neural retrievers, was contributed by \href{https://nlp.stanford.edu/}{Stanford NLP}. Since \primeqa{} provides very easy entry points into its core library, they were able to integrate their software into the retriever script \texttt{run\_ir.py} independently. Their contribution to \primeqa{} provides the QA community with the ability to obtain SOTA performance on OpenQA benchmark datasets by performing `late interaction' search on a variety of datasets. They also contributed ColBERTv2 \cite{santhanam-etal-2022-colbertv2} and its PLAID \cite{plaid} variant. The former reduces the index size by 10x over its predecessor
while the latter makes search faster by almost 7x on GPUs and 45x on CPUs.
 
\noindent \textbf{Few shot learning:} The \href{https://u.osu.edu/ihudas/}{SunLab} from \href{https://www.osu.edu/}{Ohio State University} provided the ability to easily perform few-shot learning in \primeqa{} . Their first contribution, ReasonBERT \cite{deng-etal-2021-reasonbert}, provides a pretrained methodology that augments language models with the ability to reason over long-range relations. Under the few-shot setting, ReasonBERT in \primeqa{} substantially outperforms a RoBERTa \cite{roberta} baseline on the extractive QA task. \primeqa{} gives any researcher or developer the capability to easily integrate this component in their custom search application e.g. a DPR retriever and a ReasonBERT reader.  


\noindent \textbf{Table Readers}: \href{https://en.wikipedia.org/wiki/Beihang_University}{Beihang University} and \href{https://www.microsoft.com/en-us/research/lab/microsoft-research-asia/}{Microsoft Research Asia} contributed Tapex \cite{liu2021tapex} as the first generative Table reader within \primeqa{}. Tapex proposes a novel table pre-training strategy based on neural SQL executor and achieves the SOTA on Wiki-SQL~\cite{wikisql} and Wiki-TableQuestions~\cite{wtq}.         Their contribution reused the seq2seq trainer from Transformers \cite{wolf-etal-2020-transformers} for a seamless integration into \primeqa{}. Another contribution comes from \href{https://www.lti.cs.cmu.edu/}{LTI CMU}'s \href{http://www.cs.cmu.edu/~neulab/}{NeuLab} which integrated OmniTab \cite{jiang-etal-2022-omnitab}. It proposes an efficient pre-training strategy combining natural and synthetic pre-training data. This integration happened organically as OmniTab builds on top of Tapex in \primeqa{}. Currently, their model produces the best few-shot performance on Wiki-TableQuestions, making it suitable for domain adaptation experiments for anyone using \primeqa{}.

\noindent \textbf{Custom search app for Earth Science:} Joint work between NASA and University of Alabama in Huntsville, involved creating a custom search application over scientific abstracts and papers related to Earth Science. First, using the top level scripts in \primeqa{}, they easily trained an OpenQA system on over 100k abstracts by training a ColBERT retriever and an extractive reader. Then, they were able to quickly deploy the search application using the \href{https://github.com/primeqa/create-primeqa-app}{create-primeqa-app} and make it available publicly\footnote{\href{http://primeqa.nasa-impact.net/qa}{http://primeqa.nasa-impact.net/qa}}. 

\section{Conclusion}
\label{sec:conclusion}

\primeqa{} is an open-source library designed by QA researchers and developers to easily facilitate reproduciblity and reusability of existing and future works. This is an important and valuable contribution to the community as it enables these models to be easily accessible to researchers and end-users in the rapidly progressing field of QA. Our library also provides a `service layer' that allows developers to take pre-trained \primeqa{} models and deploy them for their custom search application. \primeqa{} is built on top of the largest NLP open-source libraries and tools and provides simple python scripts as entry points to easily reuse its core components across different applications. Our easy access and reusability has already garnered significant positive traction and enables \primeqa{} to grow organically as an important resource for the rapid state-of-the-art progress within the QA community.

\bibliography{anthology,custom}
\bibliographystyle{acl_natbib}

\appendix
\section{Appendix}
\label{sec:appendix}

\begin{figure*}[htp]
  \fbox{\subfloat{}{\includegraphics[clip,trim=0 100 0 120, width=2\columnwidth]{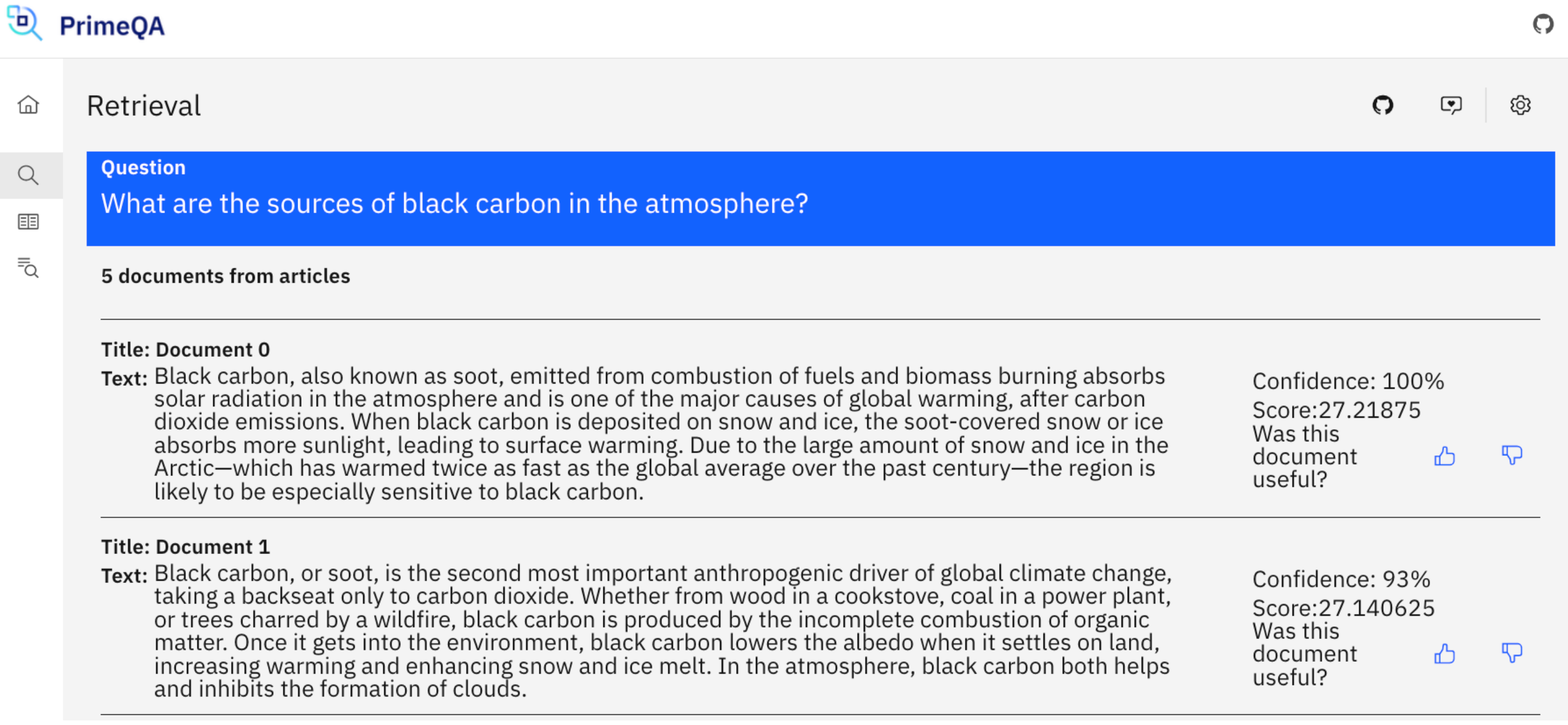}}}
  \fbox{\subfloat{}{\includegraphics[clip,trim=0 100 0 100, width=2\columnwidth]{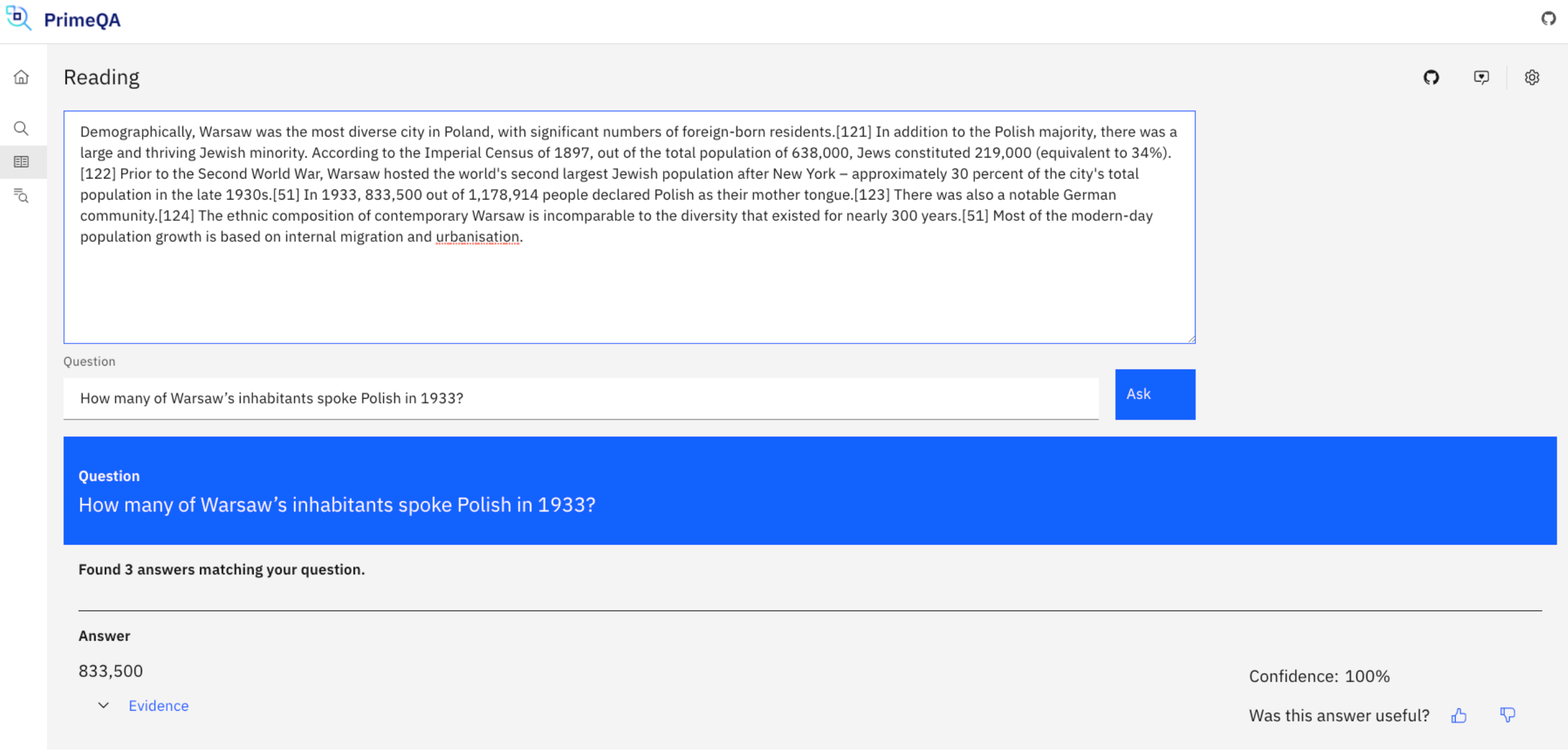}}}
  \fbox{\subfloat{}{\includegraphics[trim=0 100 0 120,clip,width=2\columnwidth]{images/primeqa_app_qa.pdf}}}
\caption{PrimeQA Applications}
\label{fig:primeqa-app-demo} 
\end{figure*}

\subsection{PrimeQA Applications}
Figure~\ref{fig:primeqa-app-demo} shows a screen-shot of three \textbf{PrimeQA} applications.

\begin{table}[t]\small		
\centering		
\begin{tabular}{ll}		
\toprule		
Datasets \\
\midrule
OpenNQ		\\
XOR-TyDi \cite{asai-etal-2021-xor}		\\
ELI5		
{SQuAD} \cite{rajpurkar-etal-2016-squad}		\\
TyDiQA \cite{tydiqa}		\\
NQ \cite{nq47761}		\\
ELI5		\\
SQA~\cite{iyyer2017search-based}		\\
WTQ~\cite{pasupat-liang-2015-compositional}		\\
DocVQA \cite{mathew2021docvqa}		\\
WikiSQL~\cite{zhongSeq2SQL2017}	\\
		
 \bottomrule		
\end{tabular}		
\caption{ A list of some of the supported datasets in PrimeQA}		
\label{tab:datasets}		
\end{table}

\begin{table}[t]\small		
\centering		
\begin{tabular}{ll}		
\toprule	
retriever & \href{https://github.com/primeqa/primeqa/blob/main/primeqa/ir/run_ir.py}{simple python script} \\
reader & \href{https://github.com/primeqa/primeqa/tree/main/primeqa/pipelines#reader-components}{inference APIs} \\
unstructured qg & \href{https://github.com/primeqa/primeqa/tree/main/notebooks/qg}{inference APIs} \\
pipeline & \href{https://github.com/primeqa/primeqa/tree/main/primeqa/pipelines#qa-pipeline}{inference APIs}. \\
		
 \bottomrule		
\end{tabular}		
\caption{ Links to PrimeQA }		
\label{tab:links}		
\end{table}	

\end{document}